\documentclass[10pt,journal,cspaper,compsoc]{IEEEtran}
\ifCLASSOPTIONcompsoc
  
\else
\fi
\ifCLASSINFOpdf
 
\else
 
\fi

\usepackage{graphicx}
\usepackage{booktabs}
\usepackage{amsmath}
\usepackage{longtable}
\newcommand{\tabincell}[2]{\begin{tabular}{@{}#1@{}}#2\end{tabular}}
\usepackage{epsfig}
\begin{document}
%
\title{A Quorum Sensing Inspired Algorithm \\for Dynamic Clustering}

\author{Feng~Tan,~\IEEEmembership{Student Member,~IEEE,}
        and~Jean-Jacques~Slotine,~\IEEEmembership{Member,~IEEE}
\IEEEcompsocitemizethanks{\IEEEcompsocthanksitem F. Tan is with Nonlinear Systems Laboratory,
        Massachusetts Institute of Technology, Cambridge, MA 02139, USA
        \protect\\
E-mail: fengtan at mit.edu
\IEEEcompsocthanksitem J.J.S. Slotine is with Nonlinear Systems Laboratory,
        Massachusetts Institute of Technology, Cambridge, MA 02139, USA
        \protect\\
E-mail: jjs at mit.edu}
\thanks{This work was sponsored in part by a grant from the Boeing Corporation}}


\IEEEcompsoctitleabstractindextext{%
\begin{abstract}
Quorum sensing is a decentralized biological process, through which a community of cells with no global awareness coordinate their functional behaviors based solely on cell-medium interactions and local decisions. This paper draws inspirations from quorum sensing and colony competition to derive a new algorithm for data clustering.

The algorithm treats each data as a single cell, and uses  knowledge of local connectivity to cluster cells into multiple colonies simultaneously. It simulates ``auto-inducers” secretion in quorum sensing to tune the influence radius for each cell. At the same time, sparsely distributed ``core cells” spread their influences to form colonies, and interactions between colonies eventually determine each cell's identity. The algorithm has the flexibility to analyze not only static but also time-varying data, which surpasses the capacity of many existing algorithms. Its stability and convergence properties are established.

The algorithm is tested on several applications, including both synthetic and real benchmarks data sets, alleles clustering, community detection, image segmentation. In particular, the algorithm's distinctive capability to deal with time-varying data allows us to experiment it on novel applications such as robotic swarms grouping and switching model identification. We believe that the algorithm's promising performance would stimulate many more exciting applications.

\end{abstract}

}

\maketitle

\IEEEdisplaynotcompsoctitleabstractindextext

\IEEEpeerreviewmaketitle

\section{Introduction}
In this paper we develop a novel clustering algorithm, inspired by
biological quorum sensin, and applicable to time-varying data. 

\begin{figure}[htbp] 
\centering 
\includegraphics[width=0.43\textwidth]{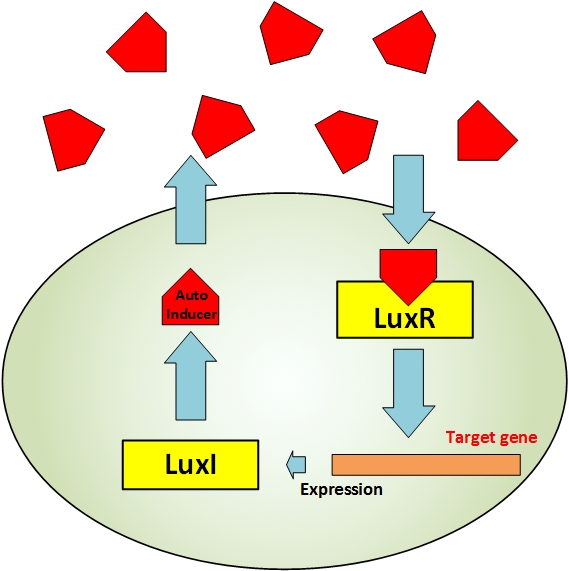}
\caption{Quorum sensing model in V. fisheri cells, adapted
  from~\cite{quo}.  Auto-inducers are secreted in the environment
  by LuxI and can be captured by receptor LuxR. The combination of
  auto-inducer and LuxR in turn stimulates more secretion of
  auto-inducers and activates gene expression for collective
  behaviors.}
\label{quorum}
\end{figure}

Quorum
sensing~\cite{quorum}~\cite{waters}~\cite{seeley}~\cite{pratt}~\cite{russo},
is a decentralized biological process by which a community of bacteria
cells interact through their local environment, with no global
information to coordinate collective behaviors. Each cell secretes
signaling molecules called auto-inducers into its environment and
builds up concentration. These auto-inducers can be captured by
receptors, which activate transcription of certain genes in the cell
(Fig. \ref{quorum}).  When few cells of the same kind exist in the
neighborhood, the density of the inducers is low, and no functional
behavior is awakened. However, when the concentration reaches a
certain threshold, a positive feedback loop is triggered to secrete
more auto-inducers and fully activate the receptors. Specific genes
start being transcribed in all cells, and functions expressed by
the genes are performed collectively. Cells can determine whether they
are surrounded in a colony by measuring the autoinducer concentration
with the response regulators. Such colony identification results in
cell clustering, and we draw inspirations from it to develop a new computational algorithm for data clustering analysis.

We propose this novel learning clustering algorithm to integrate real-time applications where data evolves over-time. We are now faced with novel and challenging control problems for groups or swarms of dynamic systems, such as manipulators, robots and basic oscilators. Especially as researches in robotics advance, traditional control theories are no longer sufficient to provide ideal solutions to all problems in the field. Although many algorithms of self-organization and group behavior work well on static data sets - we gain the toolbox of synchronization and contraction analysis from control theory and we learn the ideas of tracking and community detection from machine learning, rarely have these learning algorithms been applied to controlling real-time dynamic systems. we must acknowledge that successful applications on time-varying data in image processing, video and audio recognition have thrived in the past decades, but we believe more can be done. 

Clustering is a basic problem in data analysis and machine learning. Its task is to separate a set of unlabeled objects into clusters, so that objects in the same clusters are more similar to each other than those in other clusters. Many clustering algorithms have been developed:

\textbf{Hierarchical clustering methods} as discussed in CURE~\cite{cure}, BIRCH~\cite{birch}, have two types: one is a bottom up approach, also known as the ``Agglomerative", which starts from a state in which every single data forms its own cluster and merges small clusters as the hierarchy moves up; the other is a top down approach, also know as the ``Divisive", which starts from only one whole cluster and splits recursively as the hierarchy moves down. The hierarchical clustering algorithms intend to connect ``objects" to ``clusters" based on their distance. However, the hierarchical clustering algorithms are quite sensitive to outliers and noise. Once a misclassification happens, the algorithm is not capable of correcting the mistake in any future period.

\textbf{Centroid-based clustering methods} (K-means~\cite{kmeans}, Mean-shift~\cite{meanshift}, Medoid-shift~\cite{medoidshift}), attempt to find a centroid vector to represent a cluster, although this centroid may not be a member of the dataset. The rule to find this centroid is to optimize a certain cost function, while on the other hand, the belongings of the data are updated as the centroid is repetitively updated.

\textbf{Distribution based clustering methods} such as expectation-maximization algorithms are closely related to statistics. They tend to define clusters using some distribution candidates and gradually tune the parameters of the functions to fit the data better. However, the expectation-maximization algorithm is also very sensitive to the initial selection of parameters. It also suffers from the possibility of converging to a local optimum and the slow convergence rate.

\textbf{Density based clustering methods} as discussed in DBSCAN~\cite{DBSCAN} Mean-shift~\cite{meanshift}, Medoid-shift~\cite{medoidshift} define clusters as areas of the highest density. The low density areas are usually borders or noise region. The algorithm handles well noise and outliers in data sets because their local density is often too low to pass the threshold of clustering. However by using a predefined reachability radius, the algorithm loses its flexibility. Also, it relies on density drop to detect cluster borders, which is not easily distinguishable, and the algorithm fails to detect intrinsic cluster structures. 

\textbf{Spectral clustering methods} such as Ng-Jordan-Weiss algorithm~\cite{jordan}, Normalized
Cuts~\cite{normalized}, Power iteration clustering~\cite{pic} and Diffusion maps~\cite{diffusion}, use the spectrum(eigenvalues) of proximity matrix to perform dimensionality reduction to data sets. The algorithm has solid mathematical background, but computing eigenvalues and eigenvectors is extremely time-consuming. Furthermore, since computations in the past provide no reference for future evolution, the algorithm must be rerun entirely if the data varies with time. In contrast, the quorum sensing algorithm we develop in this paper runs continuously, tracks the variance of data and updates the result step by step.

To summarize, current techniques suffer from
several limitations: requirements of cluster number inputs;
sensitivity to outliers and noise; inability to adapt to clusters of
different density or arbitrary shapes; and most importantly the discrete and uni-direction process is hard to be integrated into real-time applications. Meanwhile, these problems are easily circumvented in nature by herds of animals - flocks of birds,
schools of fish and colonies of cells all cluster with robustness and flexibility far exceeding those of artificial algorithms. Therefore, we seek inspirations from clustering methods employed by nature to develop novel algorithms that may overcome these limitations.

Computer science has long benefited from nature-inspired theories.
Similar requirements and mechanisms may be shared by
computational science and biology, providing a basis for
developing various joint applications related e.g. to coordination,
network analysis, tracking, or vision. Biological insights can
inspire new algorithms~\cite{msb}, as in the work of Yehuda et al. on
borrowing ideas from biology to solve the maximal independent set
problem~\cite{mis}. We believe they can also help establish connections
between dynamic system control and machine learning algorithms for several reasons. First,
a biologically inspired algorithm can be designed as a dynamical
process, which suits real-time control systems. Second, biological
processes perform robustly in an environment full of disturbances, satisfying the
robustness and stability requirements for learning algorithms and
dynamic control. Third, swarms of dynamic systems are usually controlled by global decisions. Applying biological processes to the system, which are usually distributed environment such as molecules, cells, or organisms, opens the possibility of local decision making. Therefore, bridging dynamic system control with machine
learning through biology inspired algorithms is promising.

In this paper, we develop a clustering algorithm inspired by quorum
sensing and colony competition. It performs well on clustering
benchmark datasets. The unique contribution of the algorithm is its smooth integration 
with dynamic systems and control strategies. With further extensions, control
theory may be more intelligent and flexible. In the rest of the paper, we describe our algorithm in section 2. Experiment results are presented in section 3. Lastly, we discuss potential future works and
areas of extensions.
\section{Algorithm Model}
To design the algorithm, we model the process of quorum sensing and colony competitions, including the auto-inducer discretion, the local influence radius tuning, the colony establishments, interactions between colonies and colony splitting and merging processes. This chapter devotes to the making of the algorithm. We first provide a succinct overview of the entire process. The following sections explain the mathematical methods used in each stage of the simulation.

Based on the characteristics introduced above about quorum sensing, we design the algorithm as a bio-mimic process as the following:
\\a) Every single cell expands its influence by increasing an ``influence radius" in its density distribution function. Local density, which captures the total amount of influence around a particular cell, is hence maintained at a level above zero. This can be viewed as an exploration stage, when all cells reach out to check whether there exists a surrounding colony.
\\b) When the density of any cell reaches a threshold, a core cell and a
colony are established simultaneously. The newly-established colony then begins to spread its influence onto its neighboring cells and local affinity. Any influence from an established colony would also inhibit the infected cells from growing into new
core cells, hence reducing the likelihood of overlapping colonies.
\\c) Existing colonies interact with each other to minimize a cost function
to achieve optimized clustering results. In the mean time merging and splitting happen among colonies.
\\d) We obtain the clustering result by analyzing the colony vector of each
cell.

\subsection{Gaussian Distributed Density Diffusion}
We treat each data as a single cell and use the Gaussian kernel function to describe secretion of auto-inducers as:
\begin{equation}
f(\vec{x},\vec{x}_i) = e^{-\frac{\|\vec{x}-\vec{x}_i\|^2}{{\sigma_i}^2}}
\end{equation}
The Gaussian distribution defines the influence of any single cell in a local region. And $\sigma_i$ acts as the ``influence radius" measuring the secretion ability of each cell. In such way, we map all the data into a matrix $M_{n\times n}$
\begin{equation}
m_{ij} = f(\vec{x}_i,\vec{x}_j) = e^{-\frac{\|\vec{x}_i-\vec{x}_j\|^2}{{\sigma_j}^2}}
\end{equation}
As we know, $f(x,\vec{x}_i)$ is cell $i$'s influence over the environment. So $m_{ij}$ is the influence of cell $j$ on cell $i$. Moreover, $\vec{d} = M \times \vec{1}_{n\times 1}$ represents local density of each cell, where
\begin{equation}
d_i = \sum_{j \neq i}^n m_{ij} = \sum_{j \neq i}^n e^{-\frac{\|\vec{x}_i-\vec{x}_j\|^2}{{\sigma_j}^2}}
\end{equation}
If the density of a cell is high, we say this cell is ``well recognized" by its neighbors and located in a well-established colony. Also it is sensible to set a lower threshold on $m_{ij}$ to make $M$ sparse, since only local neighbors have major contributions to local density, and by doing so we can hugely save computation efforts. 

\subsection{Local Decision for Influence Radius Tuning}
Cells tune their influence radius to connect with neighbors and maintain local density. We design the process to minimize quadratic error between the density vector and a goal vector $\vec{a}=a\cdot \vec{1}_{n\times 1}$, which aims to maintain each cell's density level close to $a$. The vector $\vec{a} - \vec{d}$ is the error or the ``hunger factor". Biologically, the ``hunger factor" information is carried with the auto-inducers and captured by receptors. The minimized quadratic error function is:
$$V_{density}=\|\vec{a}-\vec{d}\|^2=\|a\cdot \vec{1}_{n\times 1} - M\cdot \vec{1}_{n\times 1}\|^2$$
To minimize it, we take the time derivative of $V_{density}$
\begin{eqnarray}
\frac{d}{dt}\|\vec{a}-\vec{d}\|^2&=&\frac{d}{dt}(\vec{a}-\vec{d})^T(\vec{a}-\vec{d}) \nonumber\\
&=&-2(\vec{a}-\vec{d})^T\frac{d}{dt}(\vec{d})\nonumber\\
&=&-2(\vec{a}-\vec{d})^T(\frac{\partial}{\partial\vec\sigma}\vec{d})\dot{\vec\sigma}
\end{eqnarray}
We name the Jacobian matrix as $J = (\frac{\partial}{\partial\vec\sigma}\vec{d})$\\
Then $$J_{ij} = \frac{2\|\vec{x}_i-\vec{x}_j\|^2}{\sigma_j^3}e^{-\frac{\|\vec{x}_i-\vec{x}_j\|^2}{\sigma_j^2}}$$
\\
\textbf{Proposition I.}
\begin{equation}
\dot{\vec\sigma} = J^T (\vec{a}-\vec{d})
\end{equation}
With this tuning policy for $\vec{\sigma}_i$'s, we have
$$
\frac{d}{dt}V_{density}=-2(\vec{a}-\vec{d})^TJJ^T(\vec{a}-\vec{d}) \leq 0\\
$$
In more details,
\begin{eqnarray}
\dot{\sigma_i} &=&\sum_{j \neq i}J_{ji}(a - d_j)\nonumber\\
&=& 2\sum_{j \neq i} \frac{\|\vec{x}_i-\vec{x}_j\|^2}{\sigma_i^3}e^{-\frac{\|\vec{x}_i-\vec{x}_j\|^2}{\sigma_i^2}} (a - d_j)
\end{eqnarray}

Here, the $(a - d_j)$ term represents the ``hunger factor" of surrounding cells, and $J_{ji}$ describes cell $i$'s potential to satisfy their needs. This proposition, however, can trigger ``over-fitting" problems. To achieve the goal that every cell's local density reaches a certain value, the algorithm may produce some ill-posed results, such as some ``super cells" with infinite influence radius, while all other cells' influence radius are reduced to 0. We improve it by adding regularization terms:
\begin{equation}
\dot{\vec{\sigma}} = J^T (\vec{a}-\vec{d}) + \beta (M - D) \vec \sigma - \alpha \vec \sigma + \vec f_{init}
\end{equation}

We regularize with terms concerning $\|\sigma\|$ and $\sum\limits_{j \neq i}^n m_{ij}\|\sigma_j-\sigma_i\|$. We add $\beta (M - D) \vec \sigma$ assuming that neighboring cells have similar secreting ability. The diagonal matrix $D$ has the entries $D_{ii} = \sum\limits_{j \neq i}^n m_{ij}$. So for cell $i$, the additional term  $\sum\limits_{j \neq i}^n m_{ij}(\sigma_j-\sigma_i)$ provides diffusive bonding between its neighboring cells' influence radius. In addition, we add inhibition term $-\alpha\vec{\sigma}$ to inhibit ``super cell" from taking shape despite of existing needs. The $\vec{f}_{init}$ term activates the exploration stage by providing initial actuation for each cell to expand its influence. The exploration stage ends when most of the cells have been recognized by their neighbors. This term acts as an activator of quorum sensing when stable interactions between cells and colonies have not yet been established. 

Adding these regularization terms also helps stabilize the system. Since $(M - D)$ is a semi-negative definite matrix, adding vectors $\beta (M - D) \vec \sigma$ and $-\alpha\vec{\sigma}$ makes the eigenvalues of the system Jacobian more negative, causing faster convergence to the equilibrium or a region close to it.

Unfortunately, this proposition is not feasible for distributed computation. In a distributed computation system, a cell requires information of all other members to make decisions. The agent-to-agent communication we describe in Proposition I. would form a complex network that makes the task of information collection of all members too ponderous to complete.
\\
\\
\textbf{Proposition II.}
\begin{equation}
\dot{\vec{\sigma}} = M (\vec{a}-\vec{d} ) + \beta (M - D) \vec \sigma - \alpha \vec \sigma + \vec f_{init}
\end{equation}
To overcome the difficulty, we replace $J^T$ with matrix $M$. Here $M (\vec{a} - \vec{d} )$ is the local hunger factor vector accumulated at the location of each cell. When local environment appears to be ``hungry" (local hunger factor is positive), a cell tends to increase its influence radius to satisfy the demand, and vice versa. The design imitates the biological process in nature, in which auto-inducers transmits the information of each cell's needs into surrounding environment following the density distribution. 

Assume that influence radius in the same colony are mostly similar, i.e. $M \approx M^T$, substituting $J^T$ with matrix $M$ does not alter the nature of the model because the entries in $J^T$ are all proportional to those in $M$ at the ratio of $\frac{\|\vec{x}_i-\vec{x}_j\|^2}{\sigma_j^3}$. \\
\\
\textbf{Proposition III.}\\
For time varying data:
\begin{eqnarray}
\frac{d}{dt}\|\vec{a}-\vec{d}\|^2&=&\frac{d}{dt}(\vec{a}-\vec{d})^T(\vec{a}-\vec{d}) \nonumber\\
&=&-2(\vec{a}-\vec{d})^T\frac{d}{dt}(\vec{d})\nonumber\\
&=&-2(\vec{a}-\vec{d})^T(J_{\sigma}\dot{\vec\sigma}+J_{X}\dot{\vec{X}})\nonumber\\
&=&-2(\vec{a}-\vec{d})^T(J_{\sigma}\dot{\vec\sigma}+\dot{\vec{d}}_{X})
\end{eqnarray}
The time derivative of the cost function has an additional term $\dot{\vec{d}}_{X}=J_{X}\dot{\vec{X}}$. Here $J_{X}$ is not a Jacobian matrix, since X is a matrix whose rows represent each data. We just use the term to represent the local density changes caused by data varying. The exact term should be:
$$\dot{d}_{Xi}=\sum_j -\frac{2}{\sigma^2_j}(\dot{\vec{x}}_j-\dot{\vec{x}}_i)^T(\vec{x}_j-\vec{x}_i)e^{-\frac{\|\vec{x}_i-\vec{x}_j\|^2}{\sigma_j^2}}$$ 
It can also be calculated as $\dot{\vec{d}}_{X}=\dot{\vec{d}}-J_{\sigma}\dot{\vec\sigma}$ as an approximation without expensive calculation.

To reduce the density fluctuation inflicted by data variation, theoretically we should have $$\dot{\vec\sigma} = J^T (\vec{a}-\vec{d}) - J_{\sigma}^{-1} J_{X} \dot{\vec{X}}$$ so that $$J_{\sigma}\dot{\vec\sigma}+J_{X}\dot{\vec{X}}=J^T (\vec{a}-\vec{d})$$  However, one problem emerges: calculating the inverse of a matrix is computationally expensive, especially when $J_{\sigma}$ is sparse and unlikely to be full rank. So we introduce an alternative way that does not erase the density fluctuation caused by motion, yet removes its influence on the cost function. In more details, by making $$\dot{\vec\sigma}=J^T (\vec{a}-\vec{d})+\dot{\vec\sigma}_X$$ we are tuning $\dot{\vec\sigma}_X$'s to make sure that $$(\vec{a}-\vec{d})^T(J_{\sigma}\dot{\vec\sigma}_X+J_{X}\dot{\vec{X}})=0$$ so that 
\begin{eqnarray}
\frac{d}{dt}\|\vec{a}-\vec{d}\|^2&=&-2(\vec{a}-\vec{d})^T(J_{\sigma}\dot{\vec\sigma}+J_{X}\dot{\vec{X}})\nonumber\\
&=&-2(\vec{a}-\vec{d})^TJJ^T(\vec{a}-\vec{d}) \leq 0\nonumber
\end{eqnarray}
In such case, we can precalculate $$\vec{l} = J_{\sigma}^T (\vec{a}-\vec{d})$$ and $\vec{k}$ where $$k_i = \dot{d}_{Xi}(a-d_i)$$ So $$(\vec{a}-\vec{d})^T(J_{\sigma}\dot{\vec\sigma}_X+J_{X}\dot{\vec{X}})=\sum_i l_i \dot{\vec\sigma}_{Xi}+\sum_i k_i $$ Note here that we cannot define $$\dot{\vec\sigma}_{Xi}=-\frac{k_i}{l_i}$$ because of two problems. First, the absolute value of $\frac{k_i}{l_i}$ can be very large or even infinite; $k_i$ measures the cost function variation caused by the motion on ``cell" $i$, so the responsibility to counteract the effects of $k_i$ should be distributed to the cell's local neighbors instead of tuning its own influence radius. So we re-designed the $\dot{\vec\sigma}_{Xi}$ tuning policy: we define $\vec{k}'=MD^{-1}\vec{k}$, where $D$ is a diagonal matrix of density vector $\vec{d}$. Moreover, the total variation of cost function caused by motion is $\sum_i k_i$. For any $k_i\le 0$, keeping it would be beneficial for cost function minimization. In such case, we can replace any $k_i< 0$ with $0$ to have a saturated and improved version of tuning policy. Thus, the tuning policy is changed to 
\begin{eqnarray}
\dot{\vec\sigma} &=& J^T (\vec{a}-\vec{d})+\dot{\vec\sigma}_X\nonumber\\
\dot{\vec\sigma}_{Xi}&=&-\frac{k'_i}{l_i}\nonumber\\
\vec{l} &=& J_{\sigma}^T (\vec{a}-\vec{d}),\nonumber\\ 
\vec{k}'&=& MD^{-1}\vec{k}\nonumber\\ 
k_i &=& \dot{ d}_{Xi}(a-d_i)\quad if\quad k_i<0\nonumber\\
\dot{\vec{d}}_{X}&=&\dot{\vec{d}}-J_{\sigma}\dot{\vec\sigma}
\end{eqnarray}
\textbf{Contraction Analysis}\\ We use contraction
analysis~\cite{contraction} to prove the convergence of both
Proposition I and II. Contraction analysis shows that, given a system
$\dot{\vec{x}}=f(\vec{x},t)$, if there exists a constant $\beta>0$,
such that for $\forall \vec{x}, \forall t \geq
0$ $$\frac{1}{2}(\frac{\partial f}{\partial \vec{x}}+ \frac{\partial
  f}{\partial \vec{x}}^T) \leq -\beta I <0$$ then all solutions
converge exponentially to a single trajectory, independent of the
initial conditions. For both propositions, we treat them as
$\dot{\vec{\sigma}}=f(\vec{\sigma})-\alpha\vec{\sigma}$, with the
Jacobian matrix $F=\frac{\partial f}{\partial \vec{\sigma}}$. For
Proposition I, after rescaling the data such that $$\forall i,j,
\|\vec{x}_i-\vec{x}_j\|^2 > a^2$$
\begin{eqnarray}
|F_{i,j}|&=&2| \frac{2\|\vec{x}_i-\vec{x}_j\|^4}{\sigma_j^6}-\frac{3\|\vec{x}_i-\vec{x}_j\|^2}{\sigma_j^4} |e^{\frac{\|\vec{x}_i-\vec{x}_j\|^2}{\sigma_j^2}} \nonumber\\
&\leq&\frac{3a}{\|\vec{x}_i-\vec{x}_j\|^2}\le \frac{3}{a} \nonumber
\end{eqnarray}
Assume after setting a threshold for $m_{ij}$'s each cell has less than $5a$ neighbors, then $|\sum_j F_{ij}|\leq 15$. Let $\alpha = 15$, we can have $\frac{\partial \dot{\vec{\sigma}}}{\partial \vec{\sigma}} = F-\alpha I$ as a negative diagonally dominant matrix, so that the system is contracting, and converging to a single equilibrium.

For Proposition II, after rescaling the data such that $$\forall i,j, \|\vec{x}_i-\vec{x}_j\| > a^2$$
\begin{eqnarray}
|F_{i,j}|&=&| \sum_j m_{ij} \frac{2\|\vec{x}_i-\vec{x}_j\|^2}{\sigma_j^3}e^{\frac{\|\vec{x}_i-\vec{x}_j\|^2}{\sigma_j^2}}| \nonumber\\
&\leq&\frac{a}{\|\vec{x}_i-\vec{x}_j\|}\le \frac{1}{a} \nonumber
\end{eqnarray}
Similarly with the less than $5a$ neighbors assumption, $|\sum_j F_{ij}|\leq 5$. Let $\alpha = 5$, we have the system for Proposition II contracting, and converging to a single equilibrium.

The convergence proof by contraction analysis here is relatively conservative. With other metrics other than the unity matrix, we may be able to achieve contraction with less stringent conditions. For example, in our simulations (shown in ), we can get the system converging to a stable equilibrium with much smaller $\alpha$ choice.

\subsection{Colony Establishments and Interactions}
In quorum sensing, when concentration surpasses a certain threshold, cells begin to produce specific functional genes to perform group behavior. We use this phenomenon as the criterion for establishing a colony. When the density of a cell $d_i$ surpasses a predefined threshold, we establish a new $j$th colony originating from it and add a $n \times 1$ colony vector $\vec{c}_j$ into the colony matrix $C$, where $C = [\vec{c_1}, \vec{c_2}, ..., \vec{c_{j-1}}]$. $\vec{c}_j$ starts with the only non-zero entry as $1$ in the $i$th term, which is also $C_{ij}$. 
\\
In the Normalized Cuts algorithm~\cite{normalized}, which is a spectral clustering algorithm widely used for image segmentation tasks, it is designed to minimize the cost function:
$$
Ncuts(A,B) = \frac{cut(A,B)}{assoc(A,V)} + \frac{cut(B,A)}{assoc(B,V)}
$$
where 
$$
cut(A,B) = \sum\limits_{i\in{A}, j\in{B}} m_{ij}$$
$$
assoc(A,V) = \sum\limits_{i\in{A}} m_{ij}
$$

Likewise, we design the colony interactions to minimize a cost function similar to the Normalized Cuts:
\begin{eqnarray}
V_{colony} &=& \sum_{i \neq j} {\vec{c}_i}^T (M + M^T) \vec{c}_j - \frac{\gamma}{2}\sum_i{\vec{c}_i}^T(M + M^T)\vec{c}_i\nonumber\\
\dot{V}_{colony} &=& \sum_{i \neq j} {\dot{\vec{c}}_i}^T (M + M^T) \vec{c}_j - \gamma\sum_i{\dot{\vec{c}}_i}^T(M + M^T)\vec{c}_i\nonumber\\
&=& \sum_i {\dot{\vec{c}}_i}^T (M + M^T) (\vec{c}_e - (\gamma + 1)\vec{c}_i)\nonumber
\end{eqnarray}
where $$\vec{c}_e=\sum_i \vec{c}_i$$ 
Here $\sum_{i \neq j} {\vec{c}_i}^T (M + M^T) \vec{c}_j$ represents the inter-colony connectivity and $\sum_i{\vec{c}_i}^T(M + M^T)\vec{c}_i$ represents the intra-colony connectivity. 

Our optimization goal is to separate the colonies so that cells in the same colony have dense connection with each other, and cells belonging to different colonies barely have connections among them, which is equivalently to minimize the cost function $V_{colony}$. Consequently, we make
$$\dot{\vec{c}}_i = -(M + M^T) \vec{c}_e + (\gamma+1)(M + M^T) \vec{c}_i $$
$$
\dot{V}_{colony} = -\sum_i  (\vec{c}_e - (\gamma + 1)\vec{c}_i)^T (M + M^T)^2 (\vec{c}_e - (\gamma + 1)\vec{c}_i)
$$
$
\ \ \ \ \ \ \ \ \   \leq 0
$

Adopting the environmental vector $\vec{c}_e$ simplifies the calculation by using global variable updates, which follows the idea of quorum sensing. We can view the interaction equations in a matrix form, where $C_e$ is a matrix with each column same as $\vec{c}_e$:
$$\dot{C} = - (M + M^T) C_e + (\gamma+1) (M + M^T) C$$

Entries in $C$ are saturated in the range of $[0, 1]$. Interactions between colonies are composed of two parts: self-expansion and mutual inhibition. When initial colonies have not been established, colony expansion pervades. It simulates a neighbor-to-neighbor infection in which cells of a colony activate their neighboring cells and pass on the colony identity. After initial colony expansion, some colonies become neighboring to each other, thus mutual inhibition comes into effect. Eventually, it reaches balance between self-expansion and inhibitions from others.

\begin{figure}[htbp] 
\centering 
\includegraphics[width=0.47\textwidth]{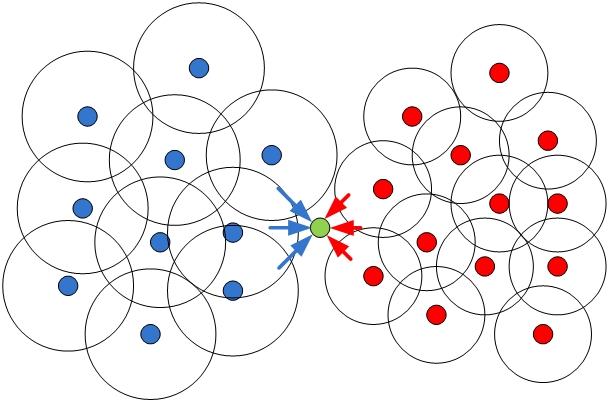}
\caption{The interactions between two colonies}
\label{gamma}
\end{figure}

Furthermore, we can illustrate such interaction in a micro view at the boundary of two competing colonies as shown in Fig. \ref{gamma}: colonies $A$ and $B$ neighboring each other, with colony vector $\vec{c}_A$ and $\vec{c}_B$, respectively. For a single cell $i$ in the boundary area, we define the interaction rules as follows:
$$\dot{c}_{Ai} = -\sum_j (m_{ij} + m_{ji}) c_{Bj} + \gamma\sum_j (m_{ij} + m_{ji}) c_{Aj}$$
$$\dot{c}_{Bi} = -\sum_j (m_{ij} + m_{ji}) c_{Aj} + \gamma\sum_j (m_{ij} + m_{ji}) c_{Bj}$$
$\gamma$ is the parameter measuring relative strength of colony inhibition and expansion. When $\gamma = 1$, we have $\dot{c}_{Ai} = -\dot{c}_{Bi}$, so if accumulated influence from colony $A$ is larger than $B$, as $$\sum_j (m_{ij} + m_{ji}) c_{Aj} > \sum_j (m_{ij} + m_{ji}) c_{Bj}$$ then finally $c_{Ai} = 1, c_{Bi} = 0$. Eventually each row in $C$ has at most one non-zero entry of $1$ that appears on the column whose colony has most accumulated influence towards the cell. When $\gamma < 1$, inhibition is enhanced, there might exist blank boundaries between colonies as the inhibition force from neighboring colonies are so strong that expanding from neither colony could reach the cell. Meanwhile when $\gamma > 1$, it is easier for colonies to spread influence into well connected neighboring colonies despite of mutual inhibition. At the beginning stage it is wise to tune up $\gamma$, to speed up newborn colonies growing, and enhance small colonies merging. Later when interactions become stable, we tune $\gamma$ back to $1$ to achieve a distinct clustering result.

\subsection{Colony Merging and Splitting}
Among the established colonies, some may be well connected to each other, while new colonies may still emerge. Such scenarios require rules for colony merging and splitting. We calculate a ratio between inter colony connections and intra colony connections measuring the possibility of merging one colony into another. We set a threshold of $r_{ij}$ for colony $i$ to be merged into colony $j$.
\begin{equation}
r_{ij} = \frac{\vec{c}_i^T (M + M^T) \vec{c}_j}{\vec{c}_i^T (M + M^T) \vec{c}_i}
\end{equation} 
On the other hand, there may be new clusters splitting from existing colonies. Since $M$ matrix is made to be sparse by setting a threshold,inter colony connections between two well segmented clusters are all zero. As a result, auto-inducers from one colony can not diffuse into other colonies. Thus, we set a continuity detecting vector $\vec{s_i}$ for each colony: evolution of $\vec{s_i}$ follows the same rules of colony interactions. When the saturated continuity detecting process reaches a stable equilibrium, we restart the entire process. Cells identified as outliers in each iteration are marked as ``not recognized" in the current round and become available for forming new colonies in next iteration.
\subsection{Clustering Result}
Finally, we get the result by choosing the maximal entry of each row in matrix $C$. The column belongs of such entries determines the colony identity of each cell. Cells with null rows are regarded as outliers.\\
The Pseudo Code of the proposed algorithm is presented below:\\
\rule{8.5cm}{0.1em}\\
1. Initialize $\vec{\sigma}$ as $\vec{0}$, form the $M$ matrix, set the parameters $a, b, \beta, \gamma$\\
2. Begin the process:\\
$\dot{\vec{\sigma}} = M (\vec{a} - \vec{d} ) + \beta (M - D) \vec \sigma - \alpha \vec{\sigma}+ \vec f_{init}$\\
Detect new cluster:\\
\indent    if $\exists d_i > b (b\leq a)$ and cell $i$ not recognized by any colony\\
\indent        create a new colony using cell $i$ as core cell\\
\indent     end\\
$\dot{C} = - (M + M^T) (C_e- C) + \gamma (M + M^T) C $\\
$\dot{S} = - (M + M^T) (S_e- S) + \gamma (M + M^T) S $\\
Cluster segmented detection:\\
\indent     if in the stable state, $S \neq C$\\
\indent     update C = S and accept new born clusters\\
\indent     end\\
$r_{ij} = \frac{\vec{c}_i^T (M + M^T) \vec{c}_j}{\vec{c}_i^T (M + M^T) \vec{c}_i}$\\
Cluster merging:\\
\indent     if $\exists r_{ij} > 0.2, i \neq j$\\
\indent     then we can merge the colony $i$ into colony $j$\\
\indent     end\\
3. Achieve the clustering results by counting the $C$ matrix\\
\rule{8.5cm}{0.1em}

For the parameters: $\gamma$ defines the ability of penetration and crossing density gaps; $\beta$ measures the similarities of $\sigma_i$'s in local neighborhood; and $a$ measures the sparsity of the connection graph. With a more connected graph, we tend to have fewer clusters. Hence, we have derived the rules for tuning the parameters: if the result suggests fewer clusters than we expect, we can tune down $a$ and $\gamma$, and if influence radius of some cells become too large, we can tune up $\beta$ and $\alpha$. 

\subsection{Comparison to Other Algorithms}
The quorum sensing algorithm we develop in this paper is not unique in applying the Gaussian affinity matrix - for example, both the density based clustering algorithms(DBSCAN, DENCLUE) and the spectral clustering algorithms( Normalized Cuts, Power Iteration Method) employ similar techniques. However, our algorithm differentiates from other clustering algorithms in many aspects. First, the density maintaining strategy we develop makes our algorithm scale-free on the cluster density issue by keeping the ``influence radius" or bandwidth adaptive for every single data. Clusters, whether dense or sparse, can all be recognized because they are not bounded by a unified bandwidth. Second, by making the bandwidth adaptive for every data, when new data flows in or drops out and the local density fluctuates, or when the data moves around which makes the cluster denser or sparser, the clustering result won't be influenced. Third, our algorithm can be extended directly to community detection applications while other algorithms cannot. Unlike the Mean-shift~\cite{meanshift} and medoid-shift~\cite{medoidshift} methods which use matrices to maximize local density as a guidance for the data to shift to a representative center, we establish a sparsely connected network with relatively constant weights between the nodes. The network then becomes the influence transmission media that determines the structure of the data and location of the clusters. The intrinsic connections in the network analysis 

We borrow the idea of normalized cuts to build a cost function that measures inter colony and intra colony connections, but we segment the clusters using different methods. The spectral clustering algorithms build on deep mathematical foundations. They analyze thoroughly the eigenvalue and eigenvector distributions of the M matrix or its modifications. The quorum sensing algorithm, as a nature-inspired algorithm, is designed heuristically by following the nature patterns. The local connectivity and diffusion are the most central properties of natural clustering among plants and animals. We therefore focus on these aspects when developing the algorithm. The spectral clustering methods, on the other hand, provide solid theoretical support to our method. 

Our estimation of the bandwidth using contraction analysis grounded in control theory is novel and sound. Various versions of clustering or density kernel estimation algorithms have been developed recently, such as Comaniciu et al's~\cite{adaptiveband}, Raykar et al's~\cite{raykar2006fast}, and Wu et al's~\cite{wu2007variable}. However, they either require complex optimization process to estimate the bandwidth, or rely on simply guiding rules that don't have solid theoretical support. In contrast, our influence radius update rules is straightforward and provides satisfying results without over-fitting.

One distinctive advantage of the quorum sensing algorithm is its ability to solve real-time dynamic problems thanks to its self-organizing natures. Current solutions for the online clustering problem are abundant: Yixin et al~\cite{yixin} develop a density-based method requiring discretization with grids, Zhong et al's~\cite{online} and Zanghi et al's~\cite{zanghi2008fast} study incremental clustering. But majority of these methods either use the grids configuration which is not plausible for high dimensional data, or stay in incremental clustering which means when new data flows in, relative clustering result can be updated in real-time. These algorithms, though work effectively on steaming data or growing databse, are not sufficient in processing time-varying data such as the locations of swarms of mobile robots or sensor information from hundreds of manipulators working on several different tasks. 

Current algorithms cannot process time-varying data because the designed processes are uni-directional, irreversible, discrete and not adaptive. They may be able to render perfect clustering results on any dataset in a single instant, yet the results soon collapse if the data change in the next millisecond; due to the discrete nature of these algorithms, history provides no valuable reference to current calculation. All computations need to be repeated to obtain updated results. In contrast, our algorithm tackles the problem. The quorum sensing algorithm possesses the characters of time-consistency, flexibility, continuity, adaptability. Our clustering result represents an integral progress where historical results can provide referring information to future result, so that calculations in the past are also utilized. Therefore, besides the satisfying performance in traditional static datasets, our algorithm is more efficient , flexible and friendly to novel applications, which will be shown in the following experiments, especially in the last two  applications.

\section{Experiments}
Our algorithm is tested on several applications, including synthetic and real benchmarks datasets, community detection, image segmentation, alleles classifications, and dynamic systems grouping and identification.

\subsection{Synthetic Benchmarks Experiments}
\begin{figure*}[htbp] 
\centering 
\includegraphics[width=0.8\textwidth]{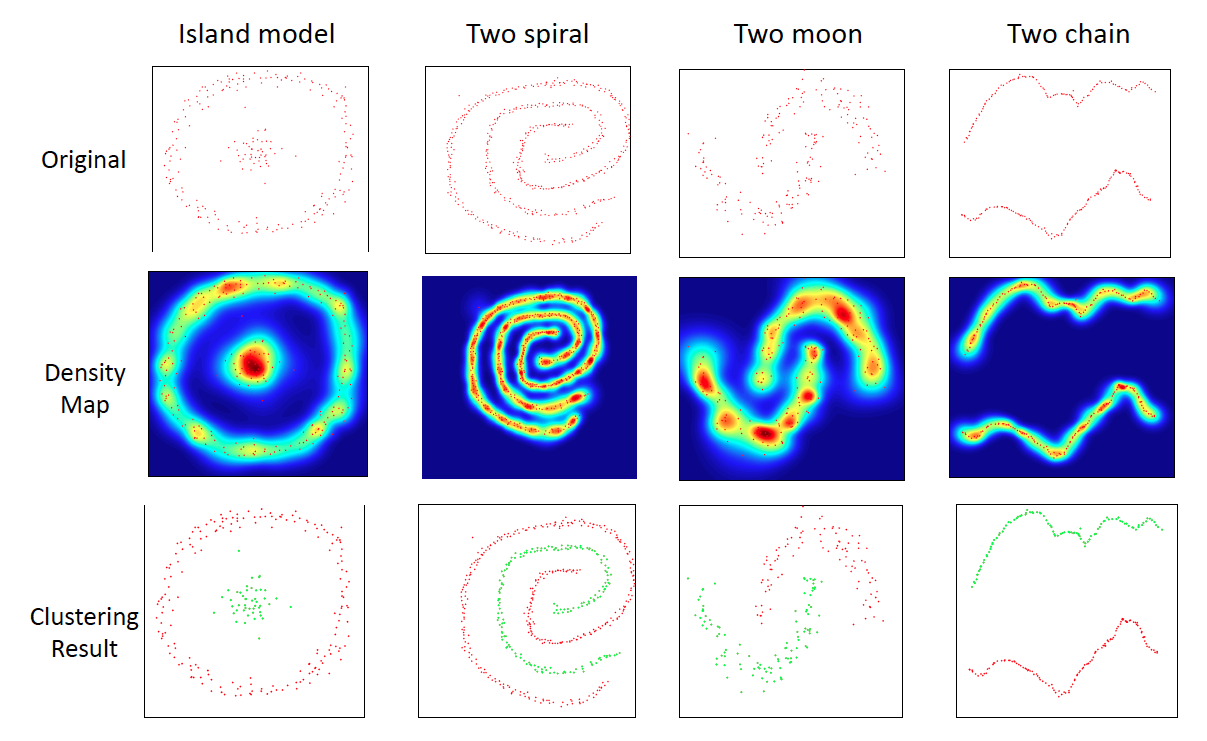}
\caption{Clustering results on synthetic benchmarks}

\label{tcdata}
\end{figure*}

We provide clustering results in Fig. \ref{tcdata} on four synthetic datasets that are only nonlinearly separable and cannot be solved by K-means or distribution based algorithms: the two-chains model(224 samples), the double-spirals model(525 samples), the two-moons model(200 samples) and the island model(236 samples). Our influence radius tuning policy ensures that the density distribution closely fits the data topology, which provides distinct separation boundaries.

\subsection{Real Benchmarks Experiments}
\noindent\textbf{Iris flower dataset}
\\
The Iris flower dataset~\cite{iris}, introduced by Sir Ronald Fisher, consists of 150 instances forming 3 clusters, of which two are only nonlinearly separable. The dataset is to quantify the morphologic variation of Iris flowers of three species(Iris setosa, Iris virginica and Iris versicolor) on four features, the length and the width of the sepals and petals.\\
\noindent\textbf{Pendigits dataset}
\\
The Pendigits datasets~\cite{pendigits} is a 16-attributes dataset of 10992 instances. The dataset is a good benchmark for testing the ability of the algorithm to cluster the data into much more than 2 clusters simultaneously in a high dimensional space. We randomly choose 1000 instances for clustering. Also, we build two subsets of the dataset, PenDigits01 with digits ``0", ``1"(easier) and PenDigits17 with digits ``1", ``7"(harder), each with 200 samples.\\
\noindent\textbf{Polbooks dataset}
\\
PolBooks~\cite{polbooks} is a co-purchase network of 105 political books. Each book is labeled ``liberal", ``conservative", or ``neutral", mostly in the first two category.
\\
We compare our results with cutting-edge algorithms including Normalized Cuts~\cite{normalized}, Ng-Jordan-Weiss algorithm~\cite{jordan} and Power Iteration Clustering~\cite{pic}, shown in Table \ref{compare}. For Iris, our performance is comparable to the cutting edge methods, we can successfully cluster
the dataset into three clusters with 4 errors out of 150 samples, a 97.3\% correctness rate. For Pen-digits, we can cluster 10 classes simultaneously with overall correctness rate 86.6\% while other methods don't have such ability. Moreover, for the two subcases, we outperform the comparisons. For the network segmentation task of Polbooks dataset, although our algorithm is not designed to solve such tasks, the result is still very satisfying. And we will discuss more in details about community detection applications in later sections.

\begin{table}[htbp]
 \caption{\label{compare}Clustering result comparison with NCut, NJW and PIC}
\centering
 \begin{tabular}{c|c|c|cccc}

  \toprule
 
  Dataset&Instances&Clusters&\tabincell{c}{NCut\\(\%)}&\tabincell{c}{NJW\\(\%)}&\tabincell{c}{PIC\\(\%)}&\tabincell{c}{Ours\\(\%)}\\
  \midrule
Iris &150&3&67.3&80.7&98.0&97.3\\
Pendigits&1000&10&&&&86.6\\
PenDigits01&200&2&100.0&100.0&100.0&100.0\\
PenDigits17&200&2&75.5&75.5&75.5&81.5\\
Polbooks&105&3&84.8&82.3&86.7&83.8\\

  \bottomrule
 \end{tabular}
\end{table}

\begin{table*}[htbp]
 \caption{\label{digits}Clustering result of Pendigits dataset}
\centering 
 \begin{tabular}{ccccccccccccc}
 
  \toprule
  Cluster&\tabincell{c}{Number\\of data}&0&1&2&3&4&5&6&7&8&9&\tabincell{c}{Hit\\rate(\%)}\\
  \midrule
  1&84&82&0 &0 &0 &0 &0 &0 &0 &2&0 &97.6\\
  2 & 81 &0&76&0&0&0&0&0&5&0&0& 93.8\\
  3 & 31 &31&0&0&0&0&0&0&0&0&0& 100.0\\
4 & 81 &0&3&0&0&0&0&78&0&0&0& 96.3\\
5 & 143 &0&25&115&0&0&0&0&2&1&0& 80.4\\
6 & 107 &0&0&0&0&102&0&0&1&0&4& 95.3\\
7 & 63 &0&0&0&0&0&0&0&63&0&0& 100.0\\
8 & 134 &0&8&0&90&0&9&0&3&6&18& 67.2\\
9 & 60 &0&0&0&0&0&7&0&0&0&53& 88.3\\
10 & 11 &0&0&0&0&0&0&0&11&0&0& 100.0\\
11 & 61 &0&0&0&0&0&61&0&0&0&0& 100.0\\
12 & 18 &0&0&0&0&0&0&0&0&0&18& 100.0\\
13 & 7 &7&0&0&0&0&0&0&0&0&0& 100.0\\
14 & 21 &0&0&0&0&0&0&0&0&21&0& 100.0\\
15 & 20 &0&0&0&0&0&0&0&0&20&0& 100.0\\
16 & 14 &0&0&0&0&0&1&0&0&0&13& 92.8\\
17 & 25 &0&0&0&0&0&0&0&0&25&0& 100.0\\
 \midrule
Overall &\tabincell{c}{961\\(39 as outliers)}&&&&&&&&&&&  86.6\\

  \bottomrule
 \end{tabular}
\end{table*}

\noindent\textbf{High-dimension Semeion Handwritten Digit dataset}
\\
The Semeion handwritten digit dataset~\cite{Semeion} was created by Tactile Srl, Brecia, Italy. It is consisted of 1593 handwritten digits stretched in a 16x16 rectangular box in a gray scale of 256 values. So this is a dataset of 1593 instances with 256 attributes, which makes it a proper dataset to test our algorithm's performance on high-dimension data and see whether the ``curse of dimensionality" will influence the overall results.   

\begin{figure*}[htbp] 
\centering 
\includegraphics[width=1\textwidth]{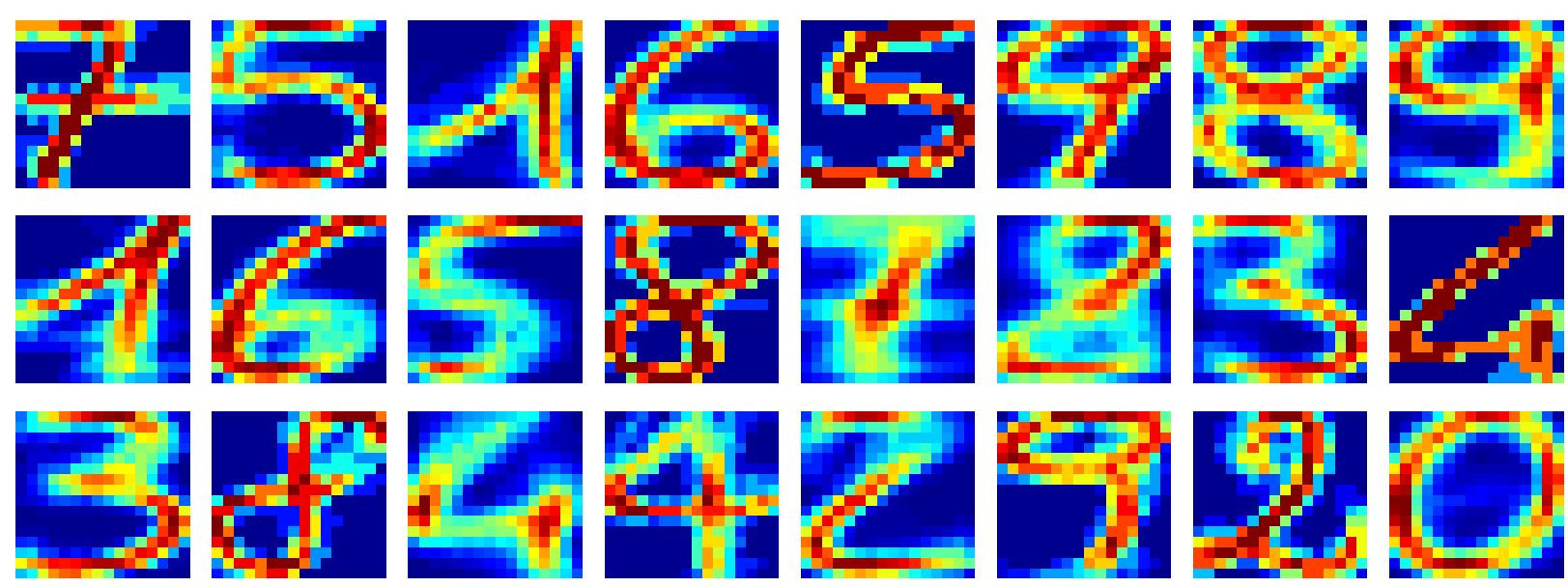}
\caption{Clustering results on Semeion Handwritten Digit dataset}

\label{Semeiondigits}
\end{figure*}

For our clustering result shown in Fig. \ref{Semeiondigits}, we have achieved 24 clusters with 76.6\% overall correctness rate. As we can see, the averaged image shown in Fig. \ref{Semeiondigits} adequately reflects the clustered shape of different digits. Detailed differentiation from writing habits for digits ``2",``8", and ``9" has been recognized and segmented in respective clusters. Among the mis-classifications, pairs of digits ``3" and ``9", ``2" and ``8", ``1" and ``7" were the major source of confusion, which is intuitive since those are also the easily misclassified pairs of hand-written digits that even sometimes confuse human readers, yet digits like ``0" and ``5" barely cause any problem.

\subsection{Experiments on Community Detection Applications}
Community detection in complex networks is currently a hot topic which attracts lots of attention in recent years. The term refers to a process that detects community structures among networked nodes and divides them into sets such that each set of nodes is densely connected internally and sparsely connected between groups. Such networks could be the internet, metabolic networks, food webs, neural networks and social networks. There are various algorithms dealing with the problem, such as minimum-cut, hierarchical clustering, Girvan–Newman algorithm, clique based methods and modularity maximization. Details about each method can be found in summarizing papers such as \cite{fortunato} and \cite{newman}. One critical advance in the area was made by Newman and Girvan~\cite{gn}, whose efforts provided a quantitative measure for the quality of a partition of a network into communities, the modularity. The modularity of a partition of a network can be written as
$$
Q=\sum_{s=1}^m[\frac{l_s}{L}-(\frac{d_s}{2L})^2]
$$
where $m$ is the number of the modules(communities) of the partition, $l_s$ is the number of links inside module $s$, $L$ is total number of links in the network and $d_s$ is the total degree of the nodes in module $s$. What the modularity measures is actually a sum up of comparison between the fraction of links inside each module with the expected fraction of links in that module if links were located at random. Though suffering from the resolution limit as introduced in \cite{limit} because of lack of local information, it is still provides a good measure of the quality of a chosen partition, especially for the modularity optimization methods that rely totally on it. Here we also use modularity for reference to compare our clustering result with a well-known fast algorithm proposed by Clauset, Newman and Moore (CNM)~\cite{cnm}, and one of its extensions proposed by Biao Xiang, En-Hong Chen, and Tao Zhou (XCZ) in ~\cite{xcz} on a American football games network Football\cite{gn} and a protein interaction network Yeast\cite{yeast}. Also we compare our results with the betweenness-based algorithm of Girvan and Newman~\cite{gn}, the fast algorithm of Clauset, Newman and Moore~\cite{cnm}, the extremal optimization algorithm of Duch and Arenas~\cite{da} and another updated algorithm of Newman~\cite{newmanpaper} on five other datasets. The networks are, in order, the karate club network of Zachary~\cite{karate}, the network of collaborations between early jazz musicians of Gleiser and Danon~\cite{jazz}, a metabolic network for the nematode Caenorhabditis elegans~\cite{metabolic}, a network of e-mail contacts at a university~\cite{email}and a trust network of mutual signing of cryptography keys~\cite{key}.

\begin{table}[htbp]
 \caption{\label{Football}Modularity compared with CNM, XCZ and CNM+XCZ}
\centering
 \begin{tabular}{c|c|c|cccc}

  \toprule
 
  Dataset&Nodes&Edges&\tabincell{c}{CNM}&\tabincell{c}{XCZ}&\tabincell{c}{CNM+XCZ}&\tabincell{c}{Ours}\\
  \midrule
Football &115&613&0.577&0.538&0.605&0.585\\
Yeast&2631&7182&0.565&0.566&0.590&0.556\\
  \bottomrule
 \end{tabular}
\end{table}  

\begin{table*}[htbp]
 \caption{\label{karate}Modularity compared with GN, CNM DA and NM}
\centering
 \begin{tabular}{c|c|c|ccccc}

  \toprule
 
  Dataset&Nodes&Edges&\tabincell{c}{GN}&\tabincell{c}{CNM}&\tabincell{c}{DA}&\tabincell{c}{NM}&\tabincell{c}{Ours}\\
  \midrule
Karate&34&156&0.401&0.381&0.419&0.419&0.404\\
Jazz&198&5484&0.405&0.439&0.445&0.442&0.419\\
Metabolic&453&4065&0.403&0.402&0.434&0.435&0.308\\
E-mail&1133&10902&0.532&0.494&0.574&0.572&0.507\\
Key singing&10680&48632&0.816&0.733&0.846&0.855&0.807\\
  \bottomrule
 \end{tabular}
\end{table*}  

With the results shown in Table \ref{Football} and Table \ref{karate}, we can find that our results applying the proposed clustering algorithm on community detection applications are comparable to top-notch algorithms in the field. Regarding the performance, we need to keep in mind that all these algorithms are modularity optimization methods that focus purely on modularity maximization, while we don't know whether they truly achieve the best partition of the network because of the limits of modularity measurement. Here we use modularity as a reference to prove our capability of solving the problem with satisfying performance.

It is an established fact that clustering and community detection are similar in many aspects: both problems share the mission of dividing dataset/network into a series of sets of data/nodes where data/nodes belonging to the same group are closer or more densely connected to each other; both problems require some measure of distance between data/nodes or clusters/groups. It is based on these shared properties that some clustering algorithms including ours can be used on network partition problems. Among these techniques, the hierarchical clustering methods are the most frequently used. The algorithms measure a distance between each pair of mini-communities and merge them all the way up. Our algorithm tackles the problem differently. We rebuild the similarity matrix while maintaining the local density at a certain level, after which we utilize the network achieved and local connectivity to determine the influence transmission.

On the other hand, critical differences between clustering and community detection exist. One key difference is the proximity measurement. For many clustering methods other than the hierarchical ones, they require a metric for distance measurement between each pair of data. In network problems, we have connectivity strength between nodes, however how to convert such information into proximity matrix that can be used by clustering algorithms creates difficult problems. In Euclidean space, if two data are far away from each other, there would be no shared neighboring data that is simultaneously close to both of them. However, in networks, influence transmission can happen easily, which makes differentiating communities using clustering methods difficult. Even two well separated nodes who respectively belongs to a densely connected community can be linked by some sparsely shared nodes, let alone in cases that some ``super-hub" nodes connecting to a large number of members in different communities. 

We propose several distance measurements.

We know that shortest path between two nodes cannot provide good measurement because shared neighboring nodes can make members from two distant communities closely connected. So we use the proximity matrix directly as the data matrix and measure distance in Euclidean space, which means $\vec{x}_i=[x_{i,1},x_{i,2},...,x{i,n}]$ where $x_{i,j}$ is the link weight between the nodes $i,j$. We define the distance between nodea $i,j$ as $\|\vec{x}_i-\vec{x}_j\|^2$ in Euclidean space. Though this metric provides good measurement on nodes connectivity, it suffers from several problems. First, it cannot deal with weighted network, since weights on the edges bring in additional differentiation other than connectivity. Second, it fails to provide enough resolution; for example when we have two nodes both disconnected from any other nodes in the network, the distance between the two nodes becomes zero , a contradiction to the assumption that they are well separated. Therefore, direct distance measurement is only one backup plan that works best in dense networks.

The most frequently used distance metric, which we adopt, is measuring the ratio between the product of the degrees of two nodes and the square of their shared connectivity as:
$$
D_{ij}=\frac{Degree\ of\ i * Degree\ of\ j}{{(Shared\ connectivity\ between\ node\ i,j)}^2}
$$
whose mathematical representation is:
$$
D_{ij}=\frac{\overrightarrow{sum(A)}*{\overrightarrow{sum(A)}}^T}{{[(A+I)^2+\epsilon]}\circ{[(A+I)^2+\epsilon}]}
$$
where $\epsilon$ is a small positive number to avoid singularity and $\overrightarrow{sum(A)}$ is a column vector calculating totaled degree of each node. In this case, we are measuring the ratio of shared neighbors between two nodes against their total degrees, such that for some sparse connections between two dense communities, when compared to the degree of the connected nodes, the sparse connections is heavily weakened because the end nodes of such edges share relatively little neighbors rather than with nodes from same community. The hub nodes problem can also be reduced. Though the hub nodes may have connections sparsely distributed in every community, if taking against their total degrees, such influence can be quickly weakened. Because they are not ``focused", one can regard these nodes as close to every community as actually far away from every community. Though this measurement is by no means perfect, our algorithm which relies on such distance definition provides satisfying clustering results on the series of network datasets presented above.

\subsection{Novel Experiment on Application for Alleles Clustering}
This section specifies the application of classifying the supertypes of the major
histocompatibility complex (MHC, also called HLA in humans) alleles,
especially the DRB (HLA-DR chain) alleles. As introduced in~\cite{smale}, it is important to understand similarities of DRB alleles for the designation of high population coverage vaccines. There are only no more than 12 HLA II alleles in each HLA gene, and an HLA gene has the ability to produce a large amount of allelic variants. It is difficult to design a vaccine that can be effective when dealing with a large population. Yet, the HLA molecules have overlapping peptide binding sets, so by grouping them into clusters, or supertypes, molecules in the same supertype will have similar peptide binding specificity. Although the Nomenclature Committee of the World Health Organization has given extensive tables on serological type assignments to DRB alleles, the information is not yet complete. So the work of grouping the alleles into groups and compare them with the WHO
labels would be helpful to the understanding of similarities of alleles and
also provide predictions to the unlabeled DRB alleles. In the work of~\cite{smale}, Wen-Jun et al. analyzed 559 DRB alleles, and proposed a kernel matrix based on BLOSUM62 matrix measuring the distance between the alleles as

$K^1 : A \times A \rightarrow R$ as
$$
K^1(x,y) = (\frac{Q(x,y)}{p(x)p(y)})^\beta, \beta>0
$$
where
$$
p(x) = \sum_{y\in A}Q(x,y), \forall x\in A
$$
Based on this, for two amino acid strings of the same length $k$, $u= (u_1, u_2, ..., u_k)$, $v= (v_1, v_2, ..., v_k)$
$$
K_k^2(u,v) = \prod_{i =1}^k K^1(u_i, v_i)
$$
and further, for two amino acid with different length $f$ and $g$
$$
K^3(f, g) = \sum_{\substack{u\subset f, v\subset g \\ \|u\|=\|v\|=k \\ all\  k=1,2,...}} K_k^2(u,v)
$$
and the normalized kernel $\hat{K}$
$$
\hat{K}(x,y) = \frac{K(x,y)}{\sqrt{K(x,y)K(y,y)}}
$$
And finally, based on the kernel matrix, for the alleles set $N$ with 559 components, the distance between the alleles is
$$
D_{L^2}(a,b) = (\frac{1}{\|N\|}\sum_{c\in N}(\hat{K}_N^3(a,c) - \hat{K}_N^3(b,c))^2)^{\frac{1}{2}}
$$  
We utilize the proposed distance in our Gaussian kernel, with the same tuning policy. The clustering result compared to~\cite{smale} is shown in Table \ref{matchrate}. Our result has no misclassifications. Yet we have classified 25 alleles as outliers, which fall into some clusters in~\cite{smale} using hierarchical clustering methods. Outliers such as DRB5*0112, DRB1*1525, DRB1*1425, DRB1*1442, DRB1*1469 and DRB1*0832 are discussed as exceptions in~\cite{smale}, which makes them more doubtful. Also we share same conclusions on exceptions like DRB1*0338, DRB3*0115; and likewise classify DRB1*1440 , DRB1*1469, DRB1*1477, DRB1*1484, DRB1*14116 and DRB1*14102 into the ST8 supertype. 
\\
Our algorithm is proved to be effective on clustering multiple clusters simultaneously for alleles data, and also our results support the conclusions of Wen-jun et al.'s work on the mathematical foundation analysis of amino acid chains. The detected outliers may lead further analysis and provide potential directions to biological researchers.
\begin{table}[htbp]
\centering
\caption{\label{matchrate}Clustering result comparison of alleles clustering}
\begin{tabular}{c|ccc}
\hline
Supertype &  Number of Alleles & Misclassified &Outliers\\
\hline
ST52 & 43 & 0 & 0\\
ST3 & 63 & 0 & 6\\
ST6 & 100 & 0 & 2\\
ST8 &52& 0& 2\\
ST4 &93 & 0 &6\\

ST2 &68 &0 &1\\
ST5 &34 & 0 & 1\\
ST53 &6& 0&0\\
ST9 &16 & 0& 1\\

ST7 &18& 0 &1\\
ST51 &15& 0 &0\\
ST1 &34&0&2\\
ST10 &3 &0 &3 \\
Overall &25&0&25\\
\hline
\end{tabular}
\end{table}

\subsection{Experiments on Image Segmentation}
Here we test our algorithm on the applications of image segmentation, which is the process of partitioning an image into multiple segments. Image segmentation can simplify the representation of an image. One very popular method of image segmentation is normalized cuts developed by Jianbo Shi and Jitendra Malik~\cite{normalized} as mentioned in previous part of the article. It is designed to minimize the cost function:
$$
Ncuts(A,B) = \frac{cut(A,B)}{assoc(A,V)} + \frac{cut(B,A)}{assoc(B,V)}
$$
where 
$$
cut(A,B) = \sum\limits_{i\in{A}, j\in{B}} m_{ij}$$
$$
assoc(A,V) = \sum\limits_{i\in{A}} m_{ij}
$$
By taking the spectral methods the second smallest eigenvector and the following eigenvectors of matrix $D^{\frac{1}{2}}(D-W)D^{\frac{1}{2}}$ provide the segmentation of the image part by part, where $D$ is the diagonal matrix of each node's degree and $W$ is the proximity matrix of the network. 

\begin{figure*}[htbp] 
\centering 
\includegraphics[width=1\textwidth]{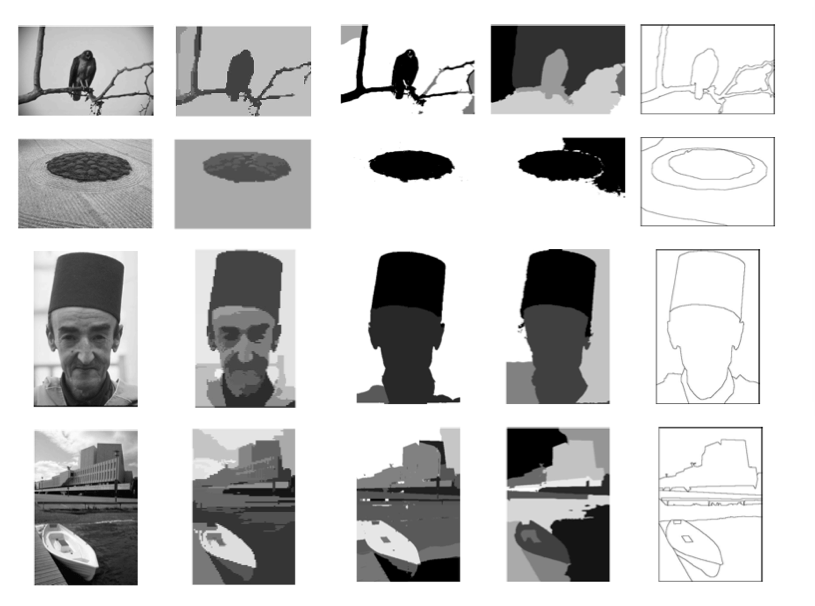}
\caption{Image segmentation results compared with normalized cuts and visual oscillators on the Berkeley dataset. First column: original images. Second: our segmentation results. Third and fourth column: results of two-layer neural oscillators and normalized cuts. Last column: human segmentation results.}

\label{image}
\end{figure*}

To apply our algorithm on image segmentation, we use both the spatial location and the intensity of pixels to formulate the distance between nodes as:
$$
D_{ij}= \|I_i-I_j\|^2*\left\{\begin{aligned}
  & \|X(i)-X(j)\|& if\ \|X(i)-X(j)\|^2<r  \\
   &0&otherwise\\
   \end{aligned}
   \right.
$$
where $X(i)$ is the spatial location of node i, and I(i) is the pixel intensity of node i. We compare our image segmentation results on selections of Berkeley benchmark images~\cite{berkeley} against results of both normalized cuts and a novel algorithm using visual oscillators~\cite{yu}. The results are shown in Fig. \ref{image}. As we can see from the results, our algorithm is fully capable of dealing with the image segmentation tasks, reflecting some details and merging connected segments with similar intensities. Based on the results, we will consider testing applications on object tracking for videos incorporating optical flows information in the future.

\subsection{Experiments on Dynamic System Grouping}
In this section, we introduce applications on clustering dynamic systems, which show our algorithm's capability on processing time varying data in a continuous way. The clustering results are flexible: they change according to the variation of data, which can be regarded as an integral over time.

\subsubsection{Application I. Real-time clustering of mobile robots}
For natural colonies such as schools of fish, flocks of birds and hordes of buffaloes, groups merging and splitting are smooth and elegant, which gives the colonies flexibility to circumvent obstacles and dangers. There are no absolute leaders or regional coordinators in the herd, yet the members coordinate and interact with local neighbors, that eventually realizes global coordination. There are already researchers working on the projects of swarm intelligence and swarm robotics, such as Vijay Kumar’s group in University of Pennsylvania and The Kilobot Project from Radhika Nagpal’s group in Harvard University. However, current researches mostly focus on the grouping behavior of tens of robots. When considering communi- cations among thousands of robot agents, we need to design a wholly novel mechanism for agents to communicate and interact with neighborhoods efficiently and quickly.

Our first application is to cluster time-varying data such as locations of mobile robots to test flexible grouping. As explained in~\cite{noise}, synchronization will enhance resistance to noise and improve robustness of network. Hence potentially, we can couple agents together to achieve synchronization by forming contracting systems as introduced in~\cite{dynamiccontraction} and~\cite{lagrangian}. Such dynamic grouping and coupling would help enhance control performance, which can be analyzed in the future.

In our simulation, 200 robots are located as the two-moon shape as previously introduced, moving around locally with radius of $0.5$ and random speed in the range of [2, 4]. During that time, 30 new robots join the group, and another subset of robots migrate to form new clusters. For potential real world application, we can use electromagnetic emitters and intensity sensors to actualize the mechanism of agent-environment-agent interaction.

\begin{figure*}[htbp] 
\centering 
\includegraphics[width=1\textwidth, height=0.7\textheight]{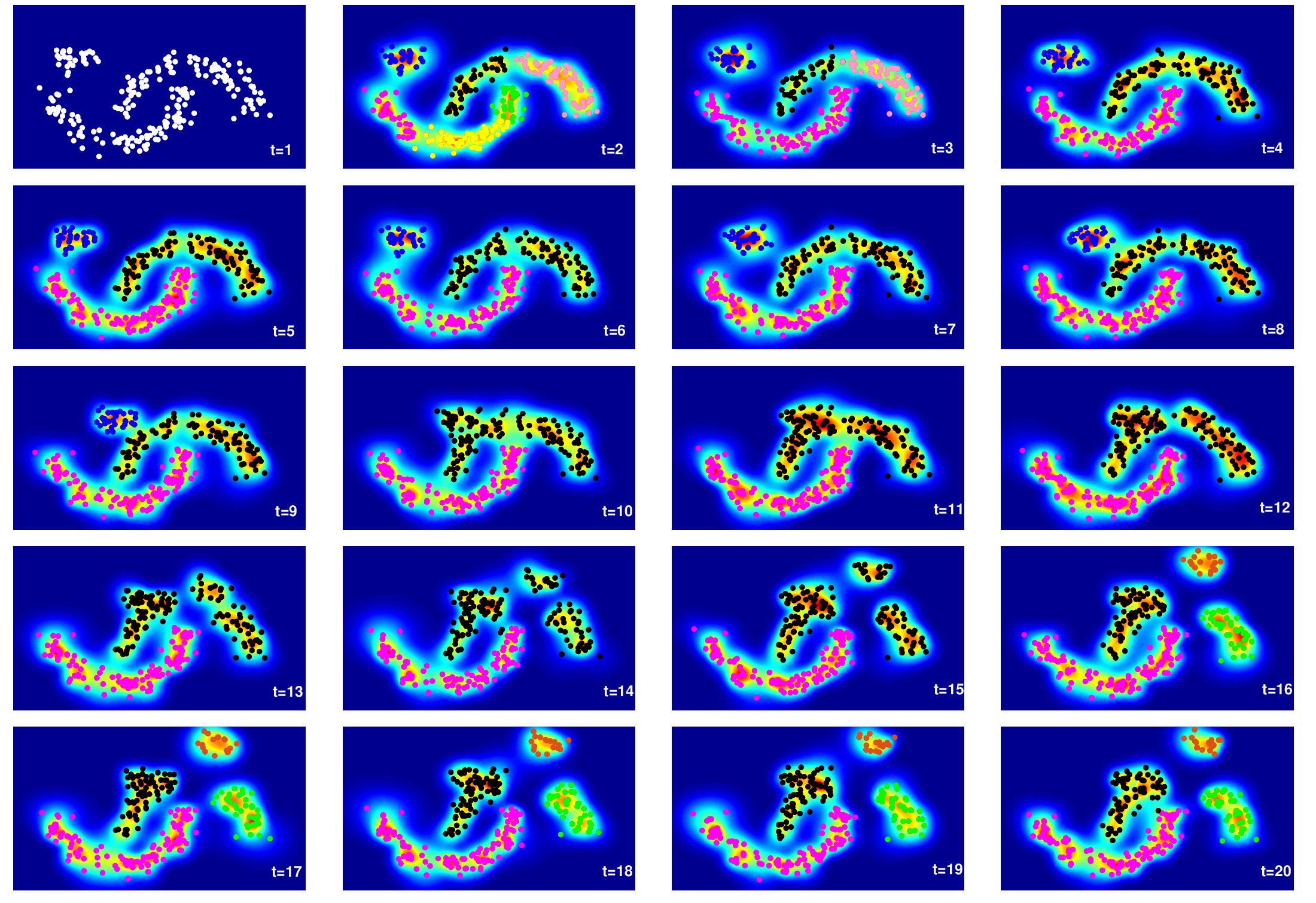}
\caption{Dynamic grouping of mobile robots locations}

\label{group}
\end{figure*}

\begin{figure}[htbp]
\centering 
\includegraphics[width=0.5\textwidth]{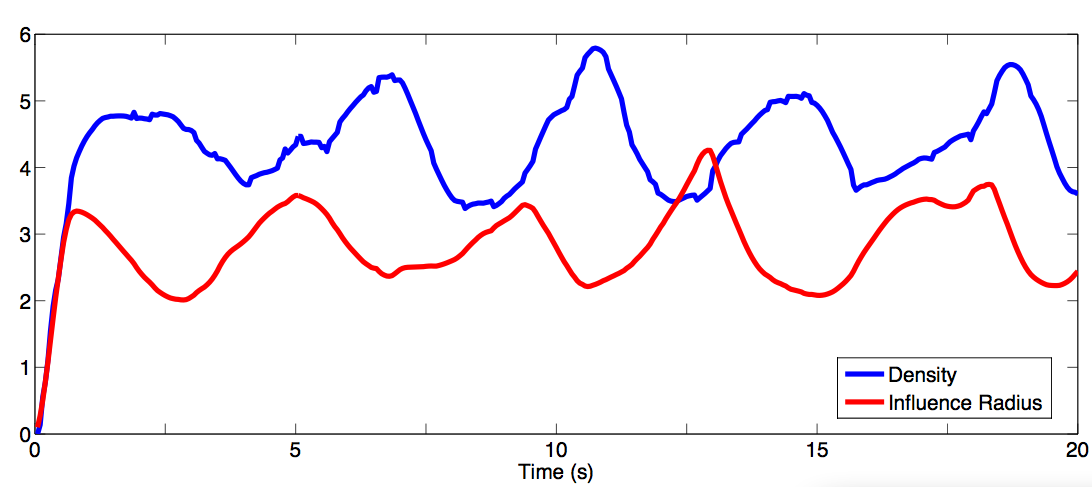}
\caption{Variation of density and influence radius of a single cell}
\label{betaY}
\end{figure}

\begin{figure}[htbp]
\centering 
\includegraphics[width=0.5\textwidth]{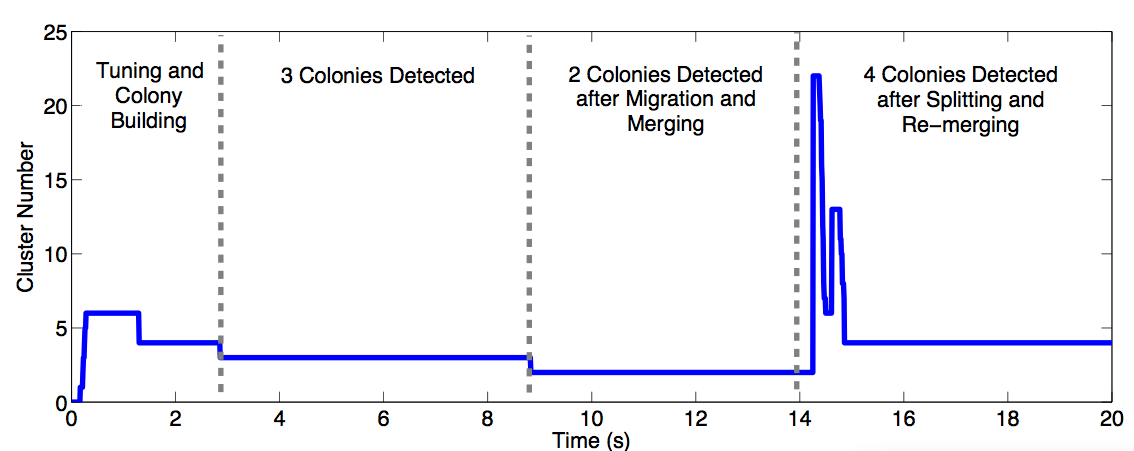}
\caption{Cluster numbers of over the simulation time}
\label{mobilecluster}
\end{figure}

From video\footnote{http://www.youtube.com/watch?v=EshxTGNpQC4}, Fig. \ref{group} and Fig. \ref{mobilecluster}, we see that the cluster number is first merged down to 3, and then varies with the merging and splitting events, exactly describing the real-time changes of robots migration. And shown in Fig. \ref{betaY}, the influence radius tuning is capable of handling local density variations: it is tuned down responding to local high density, and vice versa, to preserve balance. The results prove our capability of clustering time varying data with accumulated information, and handling variations of cluster numbers. 


\subsubsection{Application II. Multi-model switching with clustering of adaptive systems}
As introduced in~\cite{multimodel}~\cite{switch}, multi-model switching control can improve transient control performance and precision. Suppose we have many unknown dynamic systems to control, with the condition that we have the information that the parameters configurations can only fall into certain limited choices. It is possible to cluster the dynamic systems and figure out who share the same parameters configuration as one group. Here we propose a new method for multi-model switching control: 

Suppose we have a system with unknown parameters, which fall in limited possible parameters configurations.
\begin{enumerate}
\item Initially, we use adaptive control to assure acceptable performance. Simultaneously, groups of virtual systems are simulated with the same control input whose parameters scatter around the pre-known choices.
\item When the density of virtual system is stable after tuning, we calculate local density of the real system:
\begin{equation}
d_r=\sum_{i}^n e^{-\frac{\|\vec{f}_r-\vec{f}_i\|^2}{\sigma_i^2}}
\end{equation}
where $\vec{f}_r$ and $\vec{f}_i$ are Fourier transform of the input signals. 
\item If $d_r$ exceeds a predefined threshold, we know the real system belongs to a virtual cluster. Hence we can achieve the according real parameters and switch to robust control which provides more precise transient control performance.
\item Further, if the parameters vary again, by detecting $d_r$ dropping, we can resume adaptive control to maintain acceptive performance and wait for the next time that $d_r$ surpasses the threshold.
\end{enumerate}
For experiment, we use 60 virtual dynamic systems as $$m_i\ddot{x}_i+b_i|\dot{x}_i|\dot{x}_i+k_i x_i=u_i$$ $m_i,b_i,k_i$ are unknown parameters $m_i,b_i,k_i$ scattering around three known choices: $[4,3,2], [2,4,3]$ and $[3,2,4]$, with 20 systems each. And we have a ``real" system, whose parameters $m_r, b_r, k_r$ are set as [4,3,2] initially, and then changed to [2,4,3]. To track the trajectory $x_d(t)=sin(2\pi t)$, define $s=\dot{x}-\dot{x}_d + \lambda(x-x_d)$\\ and the Lyapunov function as 
\begin{equation}
V=0.5ms^2 + 0.5(\hat{m}-m)^2 + 0.5(\hat{b}-b)^2 + 0.5(\hat{k}-k)^2 \geq 0
\end{equation}
then

\begin{eqnarray*}
\dot{V}&=&m  s  \dot{s} + \dot{\hat{m}}*\tilde{m} + \dot{\hat{b}}*\tilde{b} +\dot{\hat{k}}\tilde{k}\\
&=&\dot{\hat{m}}\tilde{m}+\dot{\hat{b}}*\tilde{b}+\dot{\hat{k}}\tilde{k}+ms(\ddot{x}-\ddot{x}_d+\lambda(\dot{x}-\dot{x}_d))\\
&=&\dot{\hat{m}}\tilde{m} + \dot{\hat{b}}*\tilde{b} +\dot{\hat{k}}*\tilde{k} + s(u-m\ddot{x}_d -b|\dot{x}|\dot{x} - kx \\&&+\lambda m(\dot{x}-\dot{x}_d)) \\
\end{eqnarray*}
So by choosing control law:
\begin{equation}
 u=\hat{m} (\ddot{x}_d-\lambda (\dot{x}-\dot{x}_d)) + \hat{b} |\dot{x}|\dot{x} +\hat{k}x - k_1 s
\end{equation}
where $s=\dot{\tilde{x}}+\lambda \tilde{x}$, $k_1>0$ is a constant, and adaptation law:
\begin{equation}  \dot{\hat{m}} = - s (\ddot{x}_d-\lambda(\dot{x}-\dot{x}_d))\end{equation}
\begin{equation} \dot{\hat{b}} = - s |\dot{x}|\dot{x}\end{equation}
\begin{equation}  \dot{\hat{k}} = - sx\end{equation}

We can get $\dot{V} = - k_1s^2 \leq 0$. Also since $\ddot{V}$ is bounded, by using Barbalat's lemma, we can have  $\dot{V}$ converging to $0$ asymptotically. And thus $s$ and $x-x_d$ will converge to $0$ asymptotically since $s$ can be considered as a first order filter for $x-x_d$.

We know $$\tilde{m} (\ddot{x}_d-\lambda(\dot{x}-\dot{x}_d)) + \tilde{b} |\dot{x}|\dot{x} +\tilde{k}x$$ converges to $0$ asymptotically and thus $$u \approx {m} (\ddot{x}_d-\lambda(\dot{x}-\dot{x}_d)) + {b} |\dot{x}|\dot{x} +{k}x$$ from adaptive control theory. Systems with similar parameters require similar inputs. So that we can use the input Fourier transform vector to measure distance between systems in our algorithm in this case. 

\begin{figure}[htbp] 
\centering 
\includegraphics[width=0.5\textwidth]{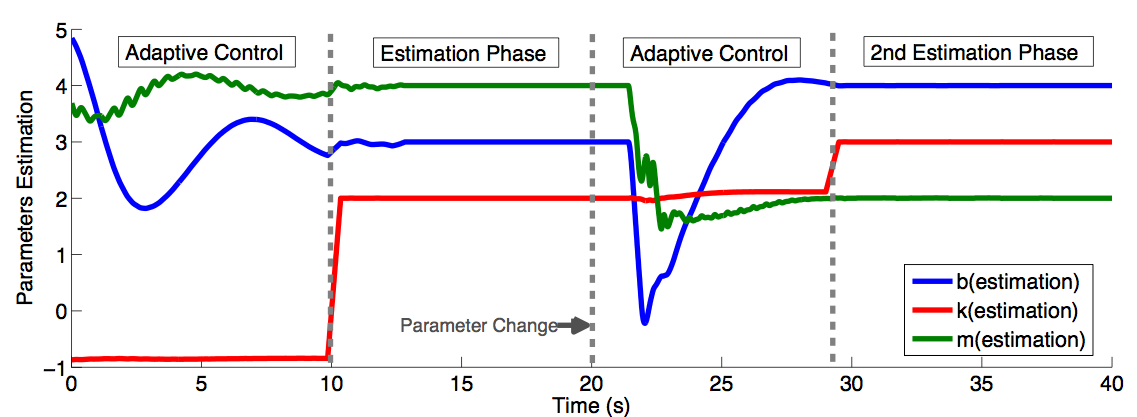}
\caption{Parameter estimations of the real system}
\label{estimation}
\end{figure}
\begin{figure}[htbp]
\centering 
\includegraphics[width=0.5\textwidth]{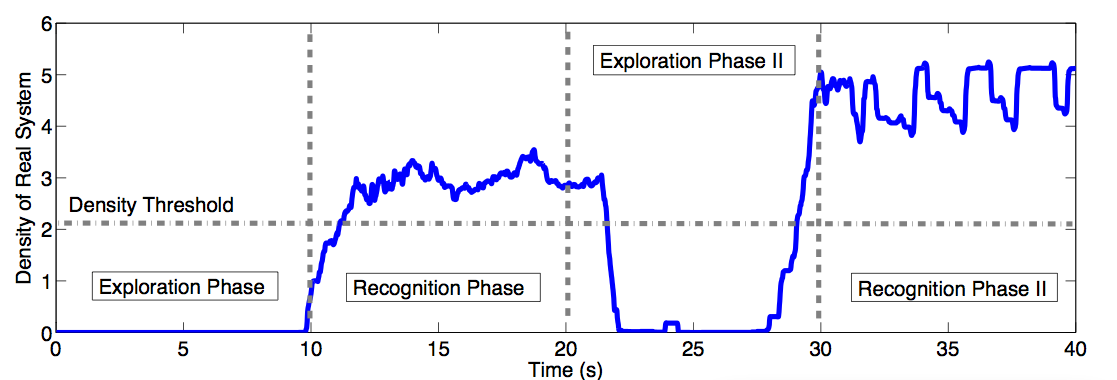}
\caption{Density of theeal system}
\label{density}
\end{figure}
Shown in Fig. \ref{estimation} and Fig. \ref{density}, soon after the multi-model switching starts at $t=10s$, density of the real system surpasses the threshold 5, and parameters are estimated correctly. After the parameters change at $t=20s$, the density drops with the control mode switched to adaptive control. After about another 10 seconds, the density is high again, and the system is correctly estimated with new parameters.

With the applications above, we show the potential of combining our algorithm with dynamic systems. The proposed algorithm imitates the smooth grouping and coordination of natural colonies and the results prove the reliability of it.

\section{CONCLUSIONS}
This paper presents a dynamic clustering algorithm inspired by quorum sensing. Our goal is to present a potential bridge between modern machine learning technologies and adaptive estimation and control. Clustering problem is chosen as a breakthrough point. We take inspirations from natural colonies that have great coordination and synchronization to search for answers and solutions. Experiments show that our algorithm performs as well as some state-of-the-art clustering methods on static datasets, and also performs well on dynamic clustering tasks with time-varying data. Extensions to applications such as community detection and image segmentation both prove to be promising. Our algorithm's advantages may be viewed as follows.
\begin{enumerate}
\item Since the influence radius is tuned to preserve local connectivity, the algorithm can adapt to clusters with different sizes and variations, even in the presence of noise and outliers. Also, it can naturally cluster data that are not linearly separable.
\item Its decentralized nature is suitable for distributed computation. For instance, robot communication may be realized with agent-environment interaction.

\item By building the adaptive density matrix or say weight evolving network, the algorithm is versatile enough to be used in community detection, image segmentation and further extending application.

\item The algorithm itself is consisted with combination of continuous dynamic systems, which is quite different from most other clustering techniques. Such differentiation endowed by the inspiration of nature contributes to the possibility of applying it to dynamic systems where other discrete algorithms are incompatible or incompetent to perform.

\item The most important merit that differentiates our algorithm from other existing ones is its highly adaptability with the varying data. The algorithm can not only adapt to new situations quickly when new data flows in, but also provide consistent clustering results on time-varying data through flexible merging and splitting. Such real-time clustering results can be used on other dynamic or real-time systems with no need of re-committing the whole process, and this is exactly what other existing algorithms cannot provide. The quorum sensing algorithm breaks free the limits of current applications of clustering and extends applications to cases like grouping swarms of robots.

\end{enumerate}

The computation complexity would be $O(n)^3$ with single processor, which means computation time is not one of our strengths. However, by making the density matrix more sparse, computation can be hugely reduced. Also, with computation power becoming more and more inexpensive, if we use the algorithm on real robots clustering, with distributed computation, the advantage of quorum sensing as a decentralized process can be fully utilized and  computation of single robot would be hugely reduced to linear time. The other potential problem of this algorithm comes from the choice of a few  parameters. During our experiments, we found that with our fixed set of parameters of radius tuning part, steady performance on various dataset is guaranteed. Yet different sets of parameters in the colony interaction part may bring different clustering results, yet they mostly fit with the distribution of data, what matters is which subgroups get merged into large ones and which not. Deeper understanding of the influence of configuration of parameters is one of our further topic.

For future work, we should study developing rules for dynamically tuning parameters, such that in different stages of algorithm, the algorithm itself can be evolving. In addition, dynamic system  metrics may require more general methods for extracting feature vectors. Last but not least, more applications involving interactions with other dynamical systems should be developed, such as video segmentation, object tracking with optical flow, and more, in particular in the contexts of synchronization and self-organizing coordination.

\addtolength{\textheight}{-10cm}   





\bibliographystyle{unsrt}
\bibliography{reference}

\end{document}